\documentclass[journal,10pt]{IEEEtran}
\usepackage[T1]{fontenc}
\usepackage[utf8]{inputenc}
\usepackage{amsmath,amssymb,amsfonts}
\usepackage{graphicx}
\usepackage[caption=false,font=footnotesize]{subfig}
\usepackage{array,booktabs}
\usepackage{placeins}
\usepackage{needspace}
\usepackage{tikz}
\usetikzlibrary{arrows.meta,positioning}
\usepackage{cite}
\usepackage{xurl}
\usepackage[hidelinks]{hyperref}
\newcommand{\papergraphic}[2][]{\includegraphics[#1]{#2}}
\newcommand{\R}{\mathbb{R}}
\hypersetup{pdftitle={Material and Defocus Cues for Occluded Object Localization in Single Hyperspectral Images: An Exploratory Study},pdfauthor={Ali Zia and Jun Zhou}}
\title{Material and Defocus Cues for Occluded Object\\
Localization in Single Hyperspectral Images:\\
An Exploratory Study}
\author{Ali~Zia and Jun~Zhou%
\thanks{Ali Zia and Jun Zhou are with the School of Information and Communication Technology, Griffith University, Queensland, Australia. Corresponding author: Ali Zia (e-mail: \href{mailto:alizia369@gmail.com}{alizia369@gmail.com}).}}
\begin{document}
\bstctlcite{occPrintAllAuthors}
\maketitle
\begin{abstract}
Partial occlusion makes object detection difficult when visible regions have weak texture or similar appearance. Hyperspectral imaging provides material-dependent spectra and wavelength-dependent optical blur, which offer complementary evidence for separating objects. We investigate a pipeline that combines nonnegative-matrix-factorization-based material grouping, divergence-based optical cues, and graph Laplacian propagation. The propagated scores support region extraction and bounding-box localization without requiring a separate depth sensor at inference. A collection of 30 indoor and six outdoor hyperspectral scenes acquired using two camera systems provides qualitative examples spanning different materials, depth arrangements, and partial occlusions. The examples illustrate how material information can support separation when optical cues are weak, and how optical cues can distinguish some surfaces with similar material properties. Objects sharing both material and depth remain challenging. This exploratory study establishes the sensing rationale and mathematical formulation, while identifying the quantitative evaluation and implementation details needed to assess detection performance systematically.
\end{abstract}
\begin{IEEEkeywords}
Hyperspectral imaging, partial occlusion, object localization, material grouping, defocus, graph Laplacian.
\end{IEEEkeywords}

\section{Introduction}\label{sec:introduction}
Partial occlusion removes contours and appearance cues needed to separate objects from one another and from the background. This difficulty affects tracking, face localization, pedestrian detection, and pose estimation~\cite{xu2013efficient,ghiasi2014occlusion,burgos2013robust,noh2018improving,Oberweger2018}. Distinctive local features can support matching on textured surfaces~\cite{Lowe2004}, but uniform surfaces and similar-looking objects provide weaker evidence. Part-based models, explicit occlusion reasoning, and temporal information address aspects of this problem~\cite{Girshick2011,Felzenszwalb2008,Meger2011,Shu2012}. Depth information offers additional geometric cues~\cite{Wang2013}, although an additional sensor introduces acquisition and registration requirements.

\begin{figure}[!t]
	\centering
	
	\subfloat[]{\papergraphic[width=0.46\linewidth]{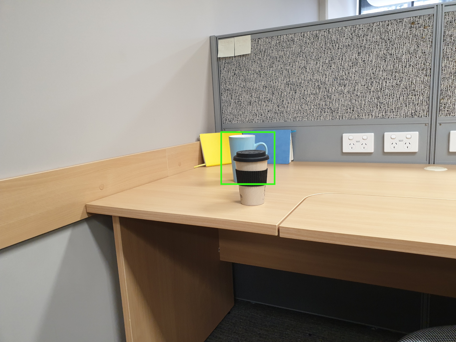}}
	~
	\subfloat[]{\papergraphic[width=0.46\linewidth]{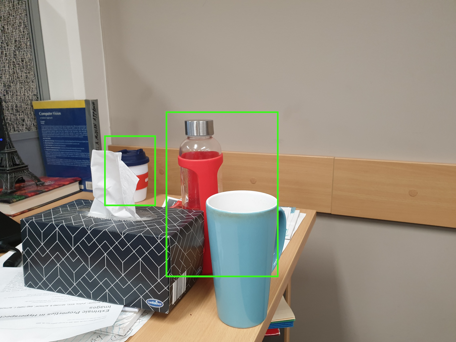}} 	
	
	\caption{Examples of partially occluded objects motivating the localization task. Visible contours and appearance alone may be ambiguous.}
\label{fig:occex}
\end{figure}

A hyperspectral image (HSI) records a spectrum at each spatial pixel. Material-dependent spectral variation can distinguish surfaces that look similar in RGB images. The same acquisition may also contain wavelength-dependent focus changes caused by chromatic aberration and defocus. These measurements motivate combining material and optical cues for localization in partially occluded scenes. Figure~\ref{fig:occex} illustrates representative ambiguities. In both examples, an occluded object retains visible evidence, but overlap interrupts its contour. The relevant question is whether the exposed region can be distinguished from adjacent surfaces. This motivates localization from complementary cues rather than reconstruction of unseen object geometry.

\subsection{Scope and Relationship to Prior Work}
This study concerns class-agnostic localization of objects that retain visible evidence in a static scene. ``Single image'' means one hyperspectral cube, potentially assembled by sequential band acquisition; it does not imply a snapshot or real-time capture. The output is a set of image regions and bounding boxes. We do not infer semantic object categories, reconstruct fully hidden surfaces, or claim amodal object extent estimation.

The spectral-depth premise builds on earlier work by Zia et al.~\cite{Zia2015,Zia2021TIP}. In particular, the 2021 journal study already integrates chromatic and defocus cues and uses material information for graph-based depth propagation. These ingredients are therefore prior work, rather than independent contributions of the present manuscript. The present focus is their use in an exploratory localization pipeline and the analysis of conditions under which object separation succeeds or fails. The study originates from the author's doctoral research~\cite{Zia2020Thesis}; the present paper provides a self-contained account of the localization study.

Related work has also used hyperspectral sensing to detect partially exposed objects in clutter. Hanson et al.~\cite{Hanson2022Occlusion} combine hyperspectral data with registered RGB-D observations for anomaly detection and robotic exposure of objects. Sun et al.~\cite{Sun2025Grasp} study a hyperspectral-guided grasping system. The present setting uses optical and material cues within one hyperspectral acquisition, without a separate depth observation at inference. This difference in sensing configuration does not establish better performance: no direct comparison is available here. Likewise, the rank-two NMF clustering strategy follows Gillis et al.~\cite{Gillis2015}; its role is to provide material-related features, not a new general-purpose clustering algorithm.

This paper describes a material--optical-cue localization pipeline, formulates its main processing stages, and examines qualitative results from indoor and outdoor scenes. The experiments establish a preliminary account of cue complementarity and associated failure modes. Section~\ref{sec:properties} introduces the cues, Section~\ref{sec:method} specifies the framework, and Sections~\ref{sec:experiments} and~\ref{sec:limitations} discuss observations and outstanding validation.

\section{Material and Optical Cues}\label{sec:properties}
\subsection{Spectral Mixing and Material Grouping}\label{subsec:blindUnmixing}
Let $\mathbf M\in\R_+^{B\times N}$ contain $B$ retained spectral bands and $N$ spatial pixels. A linear mixing approximation is
\begin{equation}\label{eq:lmm}
 \mathbf M=\mathbf Y\mathbf A+\mathbf E,
\end{equation}
where $\mathbf Y\in\R_+^{B\times K}$ contains candidate material spectra, $\mathbf A\in\R_+^{K\times N}$ contains material coefficients, and $\mathbf E$ represents noise and model error. Nonnegativity supports an NMF interpretation~\cite{Qian2011,tong2016nonnegative}. Physical abundance fractions additionally require suitable normalization and a justified mixing model. Hard spectral clustering alone does not estimate fractional abundances.

\begin{figure}[!ht]
	\centering
	\papergraphic[width=0.98\linewidth]{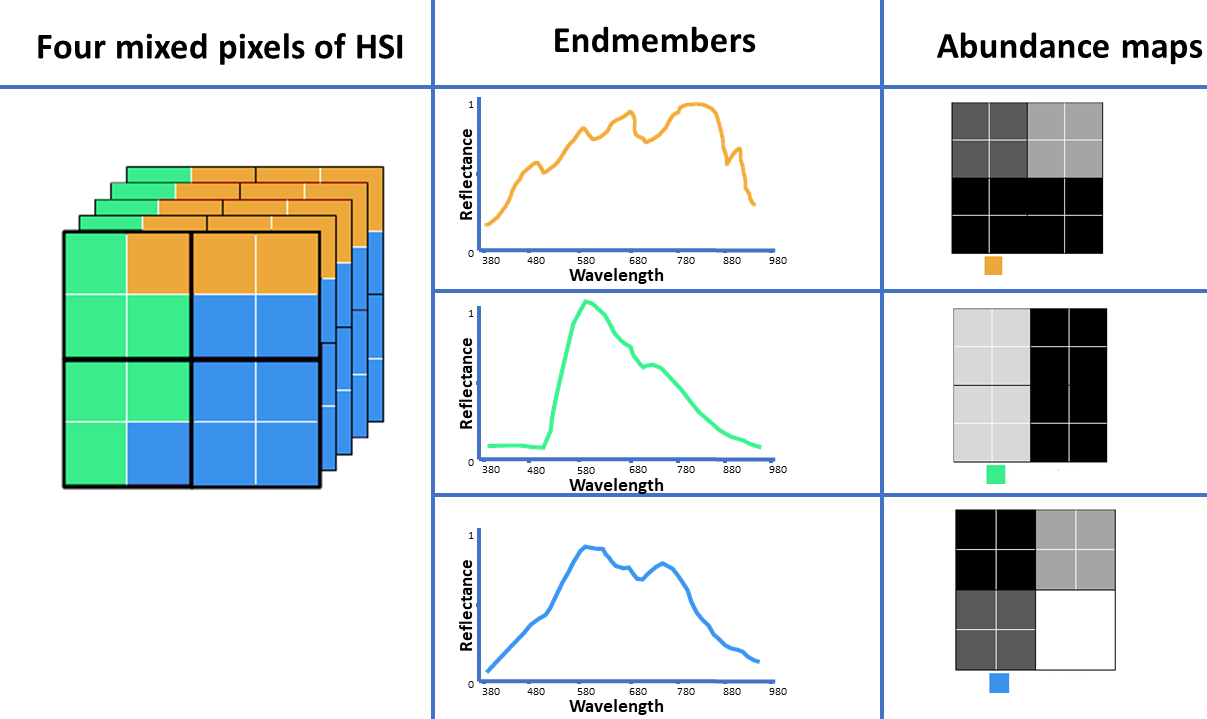}
	\caption{Illustration of the linear mixing model: mixed pixels, candidate endmember spectra, and corresponding material maps. Estimated scores need not be calibrated physical fractions.}
\label{fig:unmixing}
\end{figure}

Mixed pixels arise when several surfaces contribute to a spatial sample or when distinct materials are mixed within a surface. Figure~\ref{fig:unmixing} illustrates this interpretation. In practice, illumination, specularities, spatial blur, and nonlinear interactions may also generate components. Accordingly, the experimental images are described here as \emph{material maps} or \emph{material scores}; their interpretation as measured chemical composition is not validated. Figure~\ref{fig:unmixing} also makes an important distinction: a material component can extend across several disconnected regions, whereas an object is a spatial entity. Spectral separation must therefore be combined with spatial connectivity before assigning object regions.

\begin{figure*}[!t]
	\centering
	\subfloat[]{\papergraphic[width=0.19\textwidth]{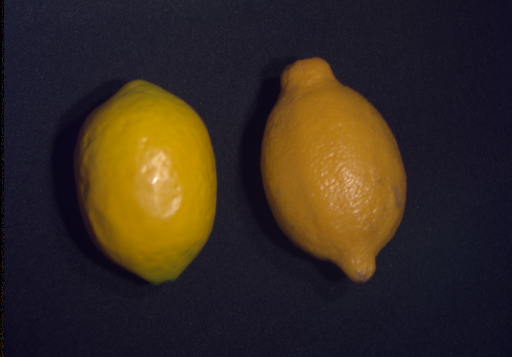}}
\hfill
	\subfloat[]{\papergraphic[width=0.19\textwidth]{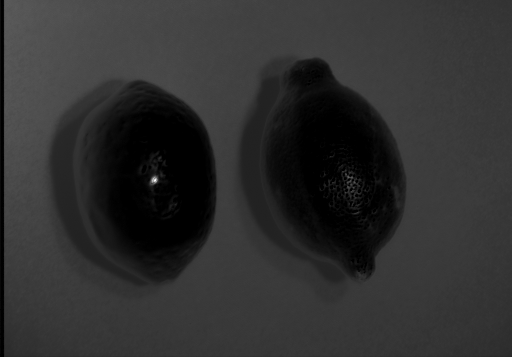}}
\hfill
	\subfloat[]{\papergraphic[width=0.19\textwidth]{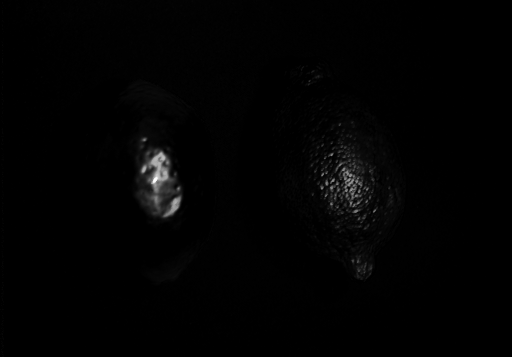}}
\hfill
	\subfloat[]{\papergraphic[width=0.19\textwidth]{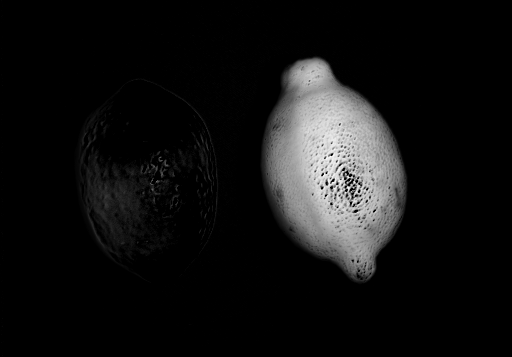}}
\hfill
	\subfloat[]{\papergraphic[width=0.19\textwidth]{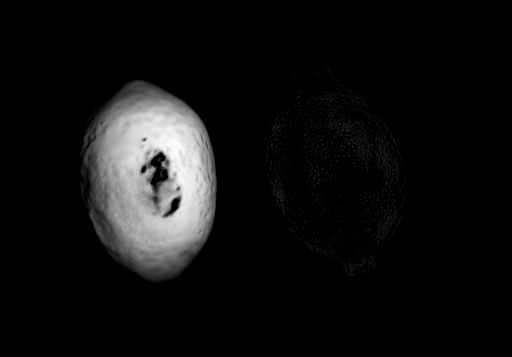}}
	\\
	\subfloat[]{\papergraphic[width=0.19\textwidth]{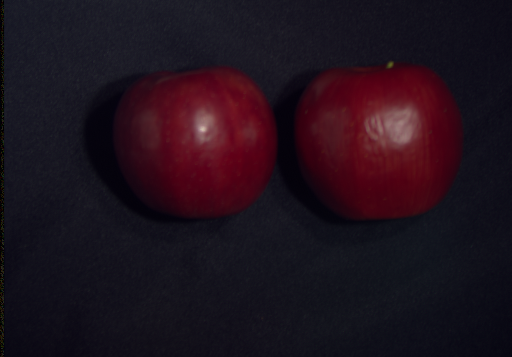}}
\hfill
	\subfloat[]{\papergraphic[width=0.19\textwidth]{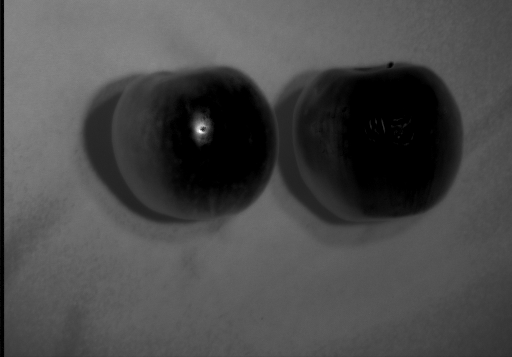}}
\hfill
	\subfloat[]{\papergraphic[width=0.19\textwidth]{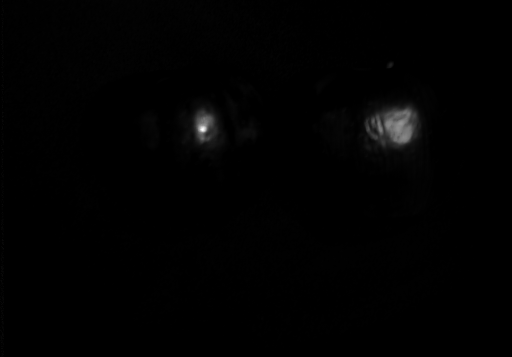}}
\hfill
	\subfloat[]{\papergraphic[width=0.19\textwidth]{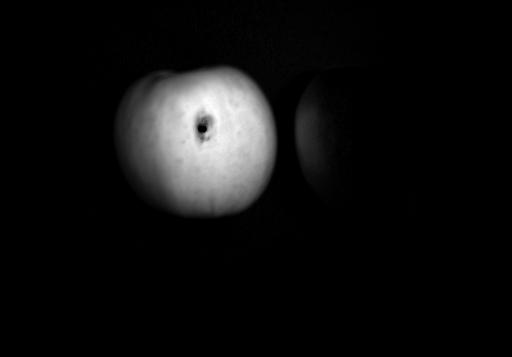}}
\hfill
	\subfloat[]{\papergraphic[width=0.19\textwidth]{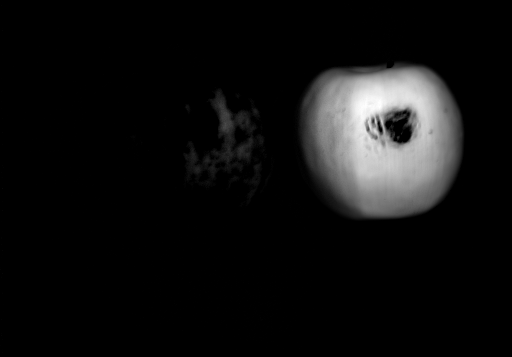}}
	
	\caption{Material-score examples for real and artificial fruit. (a), (f) RGB views; (b)--(e), (g)--(j) estimated material maps. Components can also encode background and illumination effects.}
\label{fig:abdunce maps}
\end{figure*}

Figure~\ref{fig:abdunce maps} shows examples of material maps for artificial and real fruit. The component maps respond differently to the two fruit surfaces, while some responses emphasize background or bright highlights. This suggests that spectral decomposition captures both material-related variation and imaging effects. A component should not be assigned a physical material identity solely because it isolates one object in a displayed example. The useful observation is that the maps provide contrast not directly represented by the RGB appearance; calibrated material fractions are not established.

\subsection{Chromatic Aberration and Defocus}
\begin{figure}[!ht]
	\centering
	\papergraphic[width=0.98\linewidth]{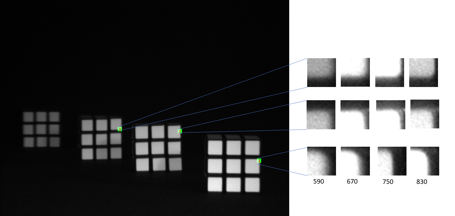}
	\caption{Wavelength-dependent focus and appearance in a hyperspectral acquisition. Wavelengths are indicated in nanometres. Spectral reflectance and optical blur both affect the observed changes.}
\label{fig:ca2}
\end{figure}

Defocus changes spatial sharpness according to scene distance and the optical configuration~\cite{pentland1987,Persch2017}. Chromatic aberration changes the focus condition with wavelength~\cite{Garcia2000,trouve2013,Zia2021TIP}. Figure~\ref{fig:ca2} illustrates wavelength-dependent appearance. Blur magnitude is not generally a globally monotonic function of depth: positions on opposite sides of the focal plane can have similar blur. Spectral reflectance, illumination, noise, and registration error also alter apparent sharpness. Relative-depth interpretation therefore requires a controlled or calibrated operating range; a local variation score alone is not a metric depth measurement. The repeated structures in Figure~\ref{fig:ca2} provide a useful visual reference for assessing edge sharpness across wavelengths. Their changing contrast must nevertheless be distinguished from changing focus. A controlled analysis would hold the scene and optics fixed and examine the band-wise edge profiles, rather than interpret intensity variation alone as a depth cue.

\section{Localization Framework}\label{sec:method}
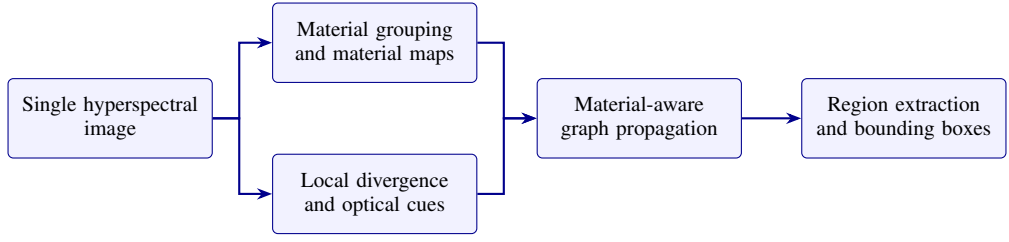
\begin{figure*}[!t]
\centering
\begin{tikzpicture}[
 font=\footnotesize,
 box/.style={draw=blue!55!black,fill=blue!5,rounded corners=2pt,
             align=center,minimum height=1.05cm,text width=2.35cm,inner sep=5pt},
 link/.style={-{Stealth[length=2mm]},thick,draw=blue!55!black}]
\node[box] (hsi) at (0,0) {Single hyperspectral\\image};
\node[box] (material) at (3.5,1.0) {Material grouping\\and material maps};
\node[box] (optical) at (3.5,-1.0) {Local divergence\\and optical cues};
\node[box] (graph) at (7,0) {Material-aware\\graph propagation};
\node[box] (regions) at (10.5,0) {Region extraction\\and bounding boxes};
\draw[link] (hsi.east) -- ++(0.35,0) |- (material.west);
\draw[link] (hsi.east) -- ++(0.35,0) |- (optical.west);
\draw[link] (material.east) -- ++(0.35,0) |- (graph.west);
\draw[link] (optical.east) -- ++(0.35,0) |- (graph.west);
\draw[link] (graph) -- (regions);
\end{tikzpicture}
\caption{Localization pipeline. Material maps define graph affinities, while local optical scores provide sparse observations. Graph propagation produces continuous scores used for region extraction and bounding-box localization.}
\label{fig:sysDiaOcc}
\end{figure*}

The pipeline extracts material features and local optical scores, propagates sparse observations on a material-aware spatial graph, and forms candidate regions. Figure~\ref{fig:sysDiaOcc} summarizes these stages. The two input branches play different roles: material features regulate which neighbouring pixels are coupled, while optical scores anchor the propagated field. This distinction explains why a visually smooth output is not sufficient evidence of correct localization. Errors in either the graph connectivity or the seed scores can remain smooth after propagation.

\subsection{Material Features from Hierarchical Clustering}\label{subsec:abundance maps}
Following the rank-two NMF framework of Gillis et al.~\cite{Gillis2015}, recursively split a set of spectral samples into material-related groups. A useful measure of departure from a single spectral direction is the best rank-one residual. For $\mathbf M_C=\mathbf M(:,C)$,
\begin{equation}\label{eq:rankone}
\begin{split}
 E(C)&=\min_{\operatorname{rank}(\mathbf X)\leq1}
       \|\mathbf M_C-\mathbf X\|_F^2\\
     &=\|\mathbf M_C\|_F^2-\sigma_1(\mathbf M_C)^2,
\end{split}
\end{equation}
where $\sigma_1$ is the largest singular value. This residual defines an approximation error; selecting the largest residual is a heuristic, not a guarantee of the greatest possible error reduction after a split. The classical separability assumption requires at least one pure pixel per material under the mixing model, not that every pixel is pure~\cite{Gillis2014,Arora2016}.

For a selected cluster $C$, obtain a rank-two nonnegative approximation
\begin{equation}\label{eq:ranktwo}
 \mathbf M_C\approx\mathbf U\mathbf H,\qquad
 \mathbf U\geq0,\quad\mathbf H\geq0,
\end{equation}
with two columns in $\mathbf U$. Fix the factor scaling by normalizing each nonzero column of $\mathbf U$ to unit $\ell_1$ norm and multiplying the corresponding row of $\mathbf H$ by its pre-normalization norm. This preserves $\mathbf U\mathbf H$ and makes the coefficient ratio independent of arbitrary NMF rescaling. A split with a zero factor column is rejected. For column $i$ of $\mathbf H$, define
\begin{equation}
 \begin{aligned}
 s_i&=\frac{H_{1i}}{H_{1i}+H_{2i}},\\
 C_1&=\{i:s_i\geq\delta\},\quad C_2=\{i:s_i<\delta\}.
 \end{aligned}
\end{equation}
Columns for which the denominator is zero carry no factorization evidence and must be assigned by an explicit implementation rule or excluded from the split criterion. Cluster indices above are local and are mapped back to the original pixel indices after splitting.

\begin{figure}[!ht]
	\centering
	\papergraphic[width=0.46\linewidth]{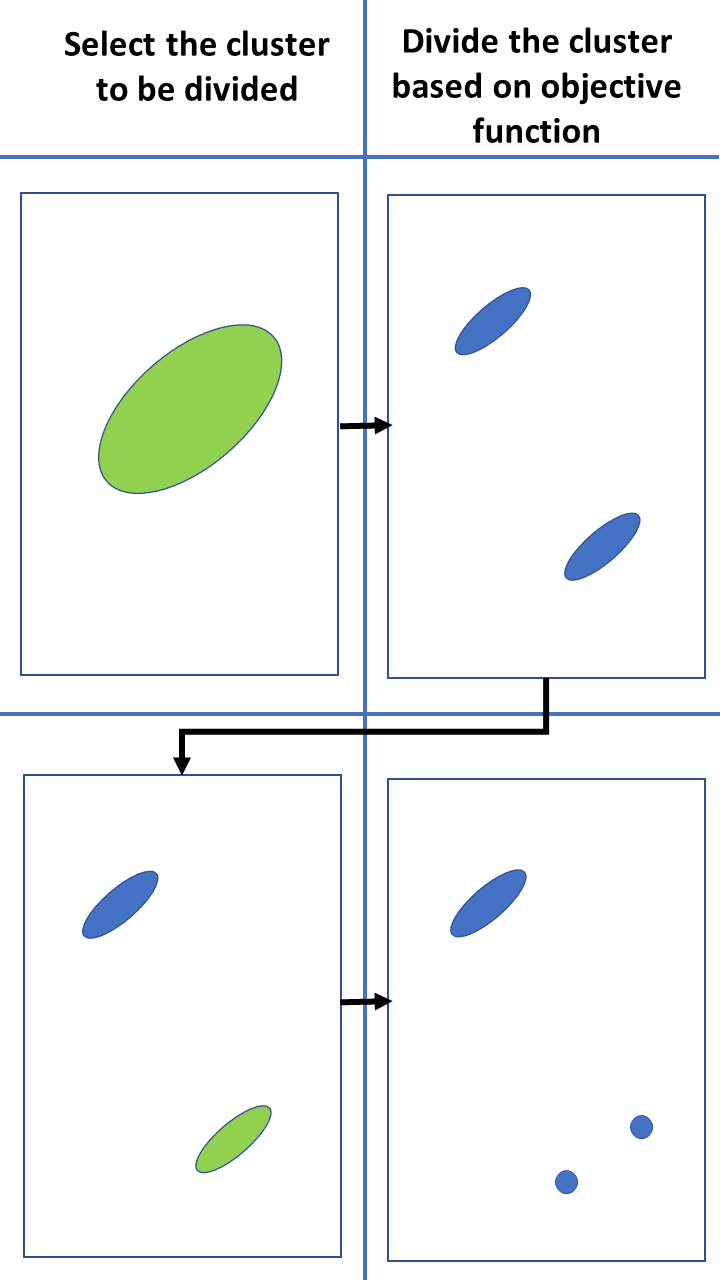}
	\caption{Schematic of divisive hierarchical clustering. The rank-two NMF clustering strategy is based on Gillis et al.~\cite{Gillis2015}.}
\label{fig:hierClus}
\end{figure}

To describe the balance--stability criterion, let $n_C$ be the number of valid columns and let $0<h<1/2$. Define
\begin{align}
 F(\delta)&=\frac{1}{n_C}\sum_i\mathbf 1\{s_i\leq\delta\},\\
 G(\delta)&=\frac{\sum_i\mathbf 1\{\delta_-\leq s_i\leq\delta_+\}}
 {n_C(\delta_+-\delta_-)},
\end{align}
where $\delta_-=\max(0,\delta-h)$ and $\delta_+=\min(1,\delta+h)$. Here $G$ is a local density estimate and need not lie in $[0,1]$. The criterion recorded in the exploratory description is
\begin{equation}\label{eq:splitcriterion}
 g(\delta)=16\bigl(F(\delta)-\tfrac12\bigr)^2+G(\delta)^2.
\end{equation}
This is a reweighted balance--stability criterion related to Gillis et al.~\cite{Gillis2015}, whose paper also discusses a quadratic alternative. It is not claimed as a separate clustering contribution. A split is admissible only if both child clusters are nonempty; if no such split exists, recursion stops. Figure~\ref{fig:hierClus} illustrates the hierarchy. Recursion permits a large spectral group to be refined, but the balance term does not explicitly protect small visible object fragments. A heavily occluded object can occupy far fewer pixels than its background. Favouring balanced child clusters may therefore conflict with extracting a rare material region; the hierarchy itself does not resolve that trade-off.

Let $\mathbf a_v\in\R_+^K$ denote the material scores at pixel $v$. A hard cluster assignment can be represented by a one-hot vector; soft maps require a separate coefficient-estimation step. The original record does not fully specify that step or its normalization. The graph formulation below accepts either representation but does not equate cluster membership with physical abundance.

\subsection{Local Divergence-Based Optical Scores}\label{subsec:KL Divergence}
\begin{figure}[!ht]
	\centering
	
	\subfloat[]{\papergraphic[width=0.46\linewidth]{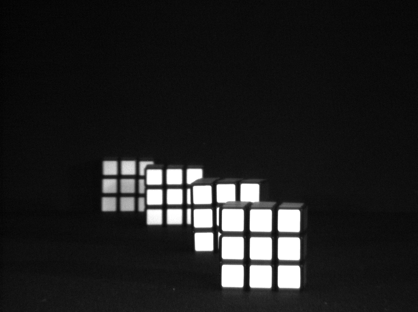}}
	~
	\subfloat[]{\papergraphic[width=0.46\linewidth]{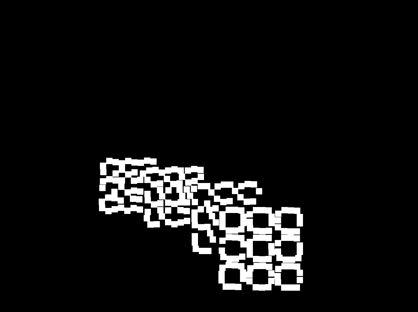}} 	
	
	\caption{(a) A band image at 700 nm. (b) Edge support used to select local neighbourhoods for optical-cue extraction.}
\label{fig:occPatches}
\end{figure}

The exploratory edge selection applies Canny detection band by band and retains locations supported by at least five bands. It then extracts nonoverlapping $t\times t$ neighbourhoods, as illustrated in Figure~\ref{fig:occPatches}. The five-band rule depends on spectral sampling, and its transfer across cameras has not been evaluated. A patch crossing a depth discontinuity may contain several depths; treating its score as one depth observation is an approximation. Figure~\ref{fig:occPatches} shows that observations are concentrated at structural edges, leaving broad uniform interiors without direct measurements. The subsequent graph stage is therefore an interpolation model, not an additional measurement source. A patch covering both an occluder and an occluded surface may mix their spectral and optical responses, so even a stable edge is not automatically a reliable single-surface seed.

\Needspace{7\baselineskip}
For a patch with $T=t^2$ pixels, define positive, normalized spectra
\begin{equation}
 p_{ib}=\frac{\max(M_{bi},0)+\epsilon}
 {\sum_{c=1}^{B}[\max(M_{ci},0)+\epsilon]},\quad\epsilon>0.
\end{equation}
This normalization removes overall spectral magnitude and therefore changes the interpretation relative to a raw-intensity score. It does not remove wavelength-dependent reflectance variation. In the limit of negligible $\epsilon$, spectra differing only by a positive intensity scale become identical after normalization. The resulting score can therefore miss brightness-only spatial contrast and is not a standalone measure of defocus or depth. For probability vectors $\mathbf p_i$ and $\mathbf p_j$, KL divergence is~\cite{Kullback1951}
\begin{equation}
 D_{\rm KL}(\mathbf p_i\|\mathbf p_j)
 =\sum_{b=1}^{B}p_{ib}\log\frac{p_{ib}}{p_{jb}}.
\end{equation}
A dimensionally explicit version of the pairwise symmetric score is
\begin{equation}\label{eq:divergence}
 d_r=\frac{1}{T(T-1)}\sum_{i\ne j}\sum_{b=1}^{B}
 \kappa_b(p_{ib}-p_{jb})\log\frac{p_{ib}}{p_{jb}},
\end{equation}
where $r$ indexes the patch, $T>1$, $\kappa_b\geq0$, and $\sum_b\kappa_b=1$. The weighted expression is nonnegative, symmetric in the pixel pair, and interpreted as a spectral-variation score. It is not itself a calibrated probability distribution or a metric depth. Band weights can emphasize a wavelength range selected by optical calibration; the exploratory account describes Gaussian-shaped weights but does not provide their fitted centre, width, or covariance adjustment. No unreported calibration procedure is assumed here.

Under controlled material and imaging conditions, blur can reduce local variation, motivating use of $d_r$ as an optical cue. Material boundaries and noise can also increase it. Assign patch scores to a sparse image vector $\boldsymbol\xi$, and separately record which pixels carry observations. Unobserved entries must not be interpreted as measured zero depth. Without a metric depth calibration, $\boldsymbol\xi$ is interpreted as an optical score map.

\subsection{Material-Aware Graph Propagation}\label{subsec:G Lap}
Graph-based propagation is established in segmentation and matting~\cite{Grady2006,Duchenne2008,Levin2008,Singaraju2009}. Construct a local undirected spatial graph with material scores $\mathbf a_v$ and pixel coordinates $\mathbf x_v$. A permutation-invariant affinity specification is
\begin{equation}\label{eq:affinity}
\begin{split}
 W_{uv}=\mathbf 1\{u\sim v\}\exp\Bigl(&-\frac{\|\mathbf a_u-\mathbf a_v\|_2^2}{2\sigma_a^2}\\
 &-\frac{\|\mathbf x_u-\mathbf x_v\|_2^2}{2\sigma_x^2}\Bigr),
\end{split}
\end{equation}
where $\sigma_a,\sigma_x>0$, $W_{vv}=0$, and $u\sim v$ is a symmetric neighbourhood relation. The widths and neighbourhood are parameters requiring explicit selection in an implementation. Using the full material vector avoids dependence on arbitrary component numbering.

Define the degree matrix and combinatorial Laplacian by
\begin{equation}
 D_{vv}=\sum_u W_{vu},\qquad \mathbf L=\mathbf D-\mathbf W.
\end{equation}
\Needspace{10\baselineskip}
For nonnegative symmetric weights, $\mathbf L$ is positive semidefinite, rather than necessarily positive definite. Let $\mathbf C=\operatorname{diag}(c_1,\ldots,c_N)$, with $c_v>0$ at observed seed locations and $c_v=0$ elsewhere. The corrected sparse-observation objective is
\begin{equation}\label{eq:propobjective}
\begin{split}
 \min_{\boldsymbol\nu}\quad
 &\frac12\boldsymbol\nu^\top\mathbf L\boldsymbol\nu\\
 &+\frac\eta2(\boldsymbol\nu-\boldsymbol\xi)^\top
 \mathbf C(\boldsymbol\nu-\boldsymbol\xi),\qquad\eta>0,
\end{split}
\end{equation}
which yields
\begin{equation}\label{eq:linearsystem}
 (\mathbf L+\eta\mathbf C)\boldsymbol\nu
 =\eta\mathbf C\boldsymbol\xi.
\end{equation}
The system has a unique solution if every connected component of the graph contains a seed with positive confidence. A component without a seed is unobserved and should be flagged or processed using a separately stated prior, rather than assigned a claimed depth estimate.

The distinction between $\mathbf C$ and $\mathbf D$ is substantive. Replacing the seed mask by the degree matrix, as in the earlier description, penalizes unobserved pixels toward the zero-filled entries of $\boldsymbol\xi$. Equation~\eqref{eq:linearsystem} corrects that modelling issue, but the original qualitative figures cannot establish its empirical effect. Moreover, a single propagated scalar does not guarantee unique object identities: objects sharing material and optical scores can remain merged.

\subsection{Regions and Bounding Boxes}
The exploratory implementation discretizes the continuous map, applies binary-image operations, and draws boxes around connected regions. Its quantization thresholds, connectivity, morphological operations, and minimum region sizes were not fully documented. We therefore do not present box extraction as a fully specified detector. A reproducible implementation must state these settings, retain spatial connectivity, and distinguish visible-region boxes from any claimed full-object extent. No semantic recognition or recovery of completely hidden objects follows from this step.

\FloatBarrier
\section{Qualitative Experiments}\label{sec:experiments}
\subsection{Scene Collection and Acquisition}
The exploratory collection contains 30 indoor and six outdoor hyperspectral scenes. Indoor scenes include toys, containers, fruit, and plants; outdoor scenes include boards, structural elements, and vegetation. All scenes were static during acquisition. The first camera combines a Brimrose acousto-optical tunable filter with a visible-to-near-infrared imaging system and records 61 bands from 400 to 1000 nm at 10 nm intervals. The second is an IMEC Snapscan VNIR system described as supporting up to 150 bands over 470--900 nm. These are acquisition-system descriptions, not a complete per-scene record of retained bands.

The original account reports removing low-signal bands near the ends of the first system's range. It does not report the exact retained indices for each scene, exposure settings, patch size, complete optical calibration, material-map estimation settings, or camera-stratified results. This limits reproducibility and prevents a claim of validated cross-camera transfer. The collection is described as a study dataset; no public download or annotation release is claimed.

\subsection{Evidence and Evaluation Scope}
The following figures are qualitative observations from the original exploratory runs. There are no reported ground-truth instance masks, standardized occlusion levels, detection AP, precision--recall curves, depth errors, runtime measurements, or baseline comparisons. Consequently, the examples cannot establish superiority, statistical significance, or a quantitative gain from cue fusion.

\subsection{Different Materials and Depth Arrangements}
\begin{figure*}[!t]
\centering\setlength{\tabcolsep}{2pt}
	
	\begin{tabular}{ccccccc}
		(a) & (b) & (c) & (d)  & (e)  & (f)  & (g) \\
		\papergraphic[width=0.132\textwidth]{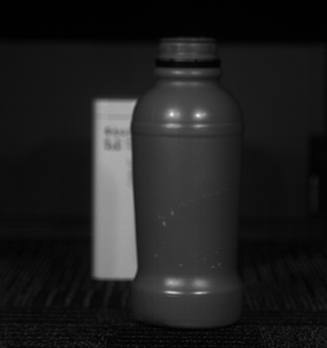}&
		\papergraphic[width=0.132\textwidth]{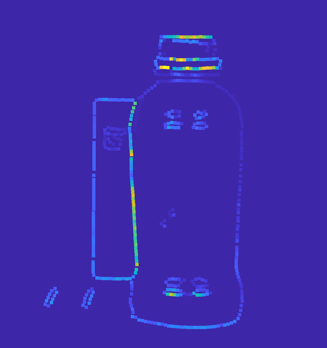}&
		\papergraphic[width=0.132\textwidth]{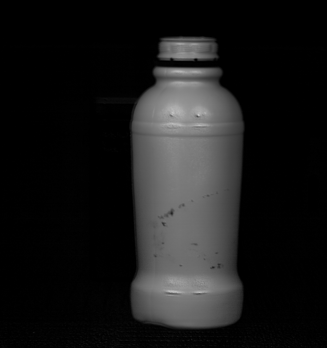}&
		\papergraphic[width=0.132\textwidth]{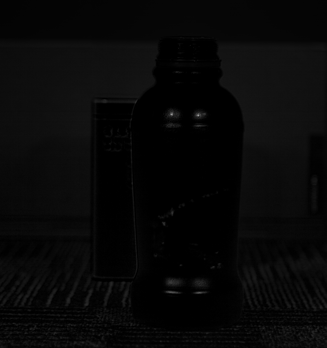}&
		\papergraphic[width=0.132\textwidth]{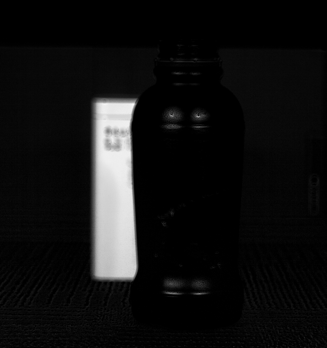}&
		\papergraphic[width=0.132\textwidth]{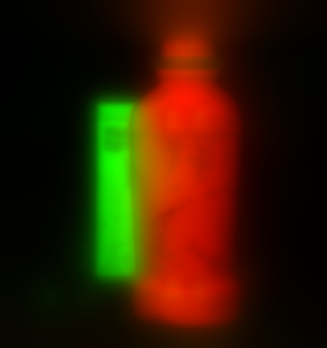}&
		\papergraphic[width=0.132\textwidth]{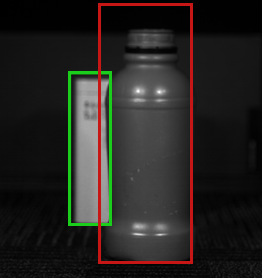}\\
		
		\papergraphic[width=0.132\textwidth]{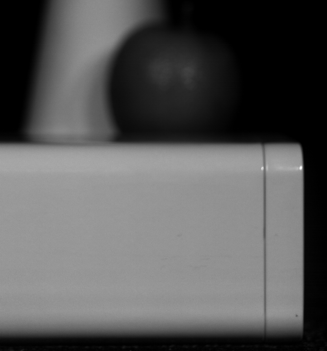}&
		\papergraphic[width=0.132\textwidth]{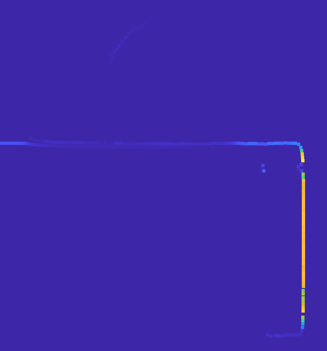}&
		\papergraphic[width=0.132\textwidth]{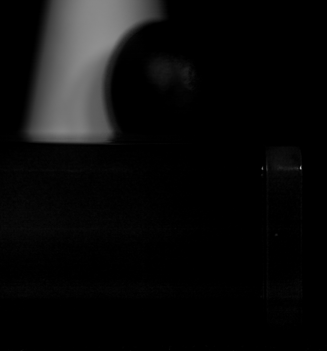}&
		\papergraphic[width=0.132\textwidth]{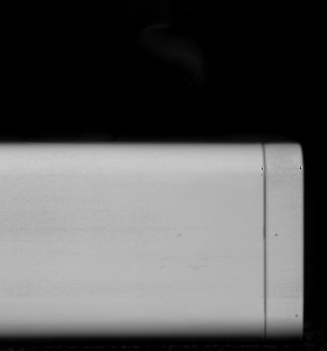}&
		\papergraphic[width=0.132\textwidth]{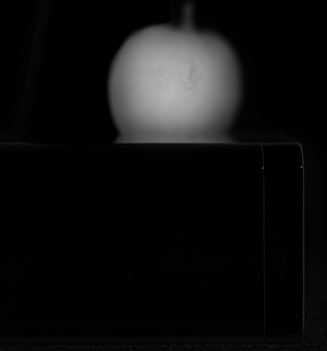}&
		\papergraphic[width=0.132\textwidth]{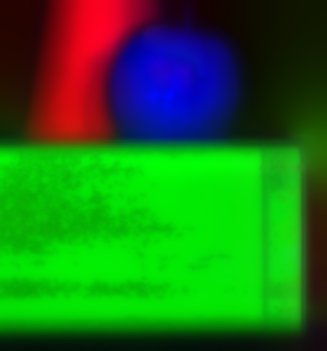}&
		\papergraphic[width=0.132\textwidth]{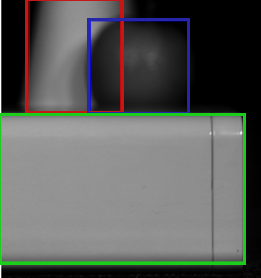}\\
		
		\papergraphic[width=0.132\textwidth]{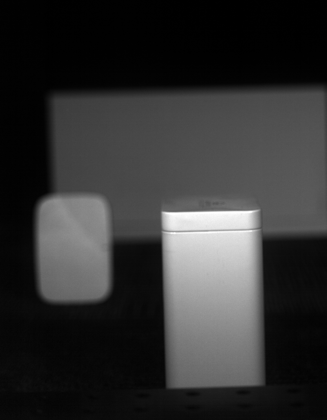}&
		\papergraphic[width=0.132\textwidth]{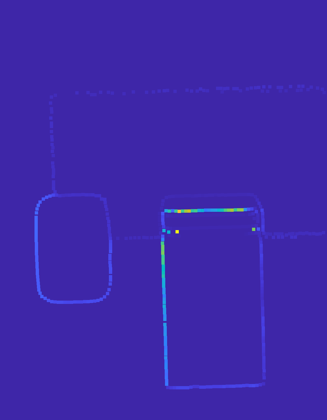}&
		\papergraphic[width=0.132\textwidth]{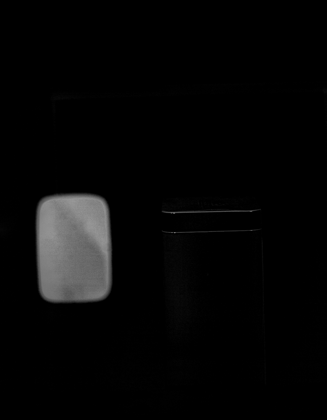}&
		\papergraphic[width=0.132\textwidth]{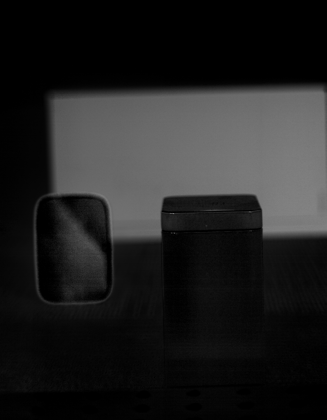}&
		\papergraphic[width=0.132\textwidth]{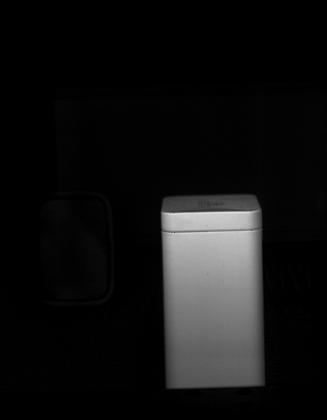}&
		\papergraphic[width=0.132\textwidth]{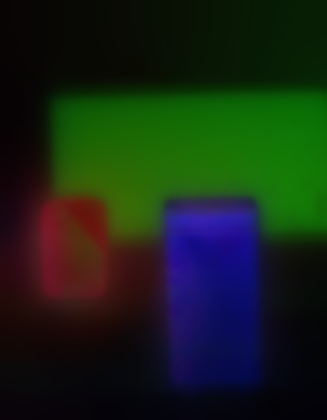}&
		\papergraphic[width=0.132\textwidth]{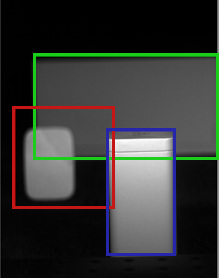}
		
	\end{tabular}
	
	\caption{Qualitative results from the original exploratory implementation. (a) Band image at 770 nm; (b) sparse optical cue map; (c)--(e) material maps; (f) continuous graph-propagated scores; (g) resulting bounding boxes. Colours visualize scores, not semantic classes or metric depth.}
\label{fig:ocres1} 
\end{figure*}

Figure~\ref{fig:ocres1} illustrates three indoor arrangements and separates the information contributed by each stage. The band images in column (a) contain the visible scene structure; column (b) contains sparse optical responses; columns (c)--(e) show selected material maps; and columns (f)--(g) show propagated scores and region boxes. The comparison should be read across each row, since colours in the score maps are visualization values rather than persistent object identities across scenes.

In the first row, the plastic bottle partly occludes a metal container. Selected material maps respond to different exposed surfaces, while the sparse optical map mainly supports boundaries. The dense map fills regions that have few direct observations. This is consistent with the intended role of material-aware propagation: retaining spectral separation while extending sparse information spatially. The bounding box around the partly visible container does not demonstrate recovery of its hidden shape or true full-object extent.

The second row is informative because the optical map provides limited interior evidence, yet material maps distinguish several visible regions. The final map therefore cannot be interpreted as a visualization of independently measured depth at every pixel. It combines local optical evidence with the spatial structure of the spectral grouping. The scene motivates cue complementarity, but does not isolate how much each branch contributes without a material-only and optical-only comparison.

The third row includes wood and cardboard, whose related spectral behaviour can reduce material contrast. Its propagated map contains broad transitions, and the region boxes are sensitive to how the continuous values are discretized. This example exposes the distinction between material separation and instance localization: one material response can cover several surfaces, while one object can generate several responses. The qualitative boxes establish an output of the pipeline, not a measured localization accuracy.

\subsection{Similar Depth: Different or Similar Materials}
\begin{figure*}[!t]
	\centering	
	\subfloat[]{\papergraphic[width=0.135\textwidth]{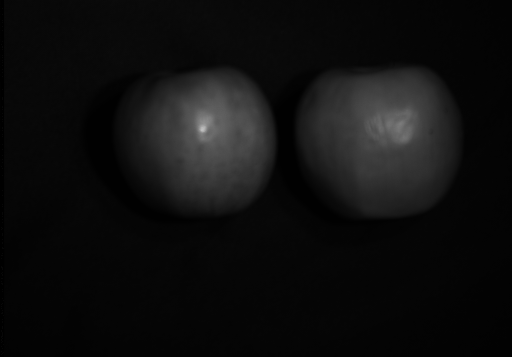}}
\hfill
	\subfloat[]{\papergraphic[width=0.135\textwidth]{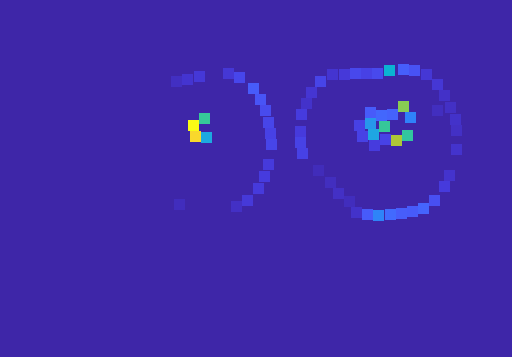}}
\hfill
	\subfloat[]{\papergraphic[width=0.135\textwidth]{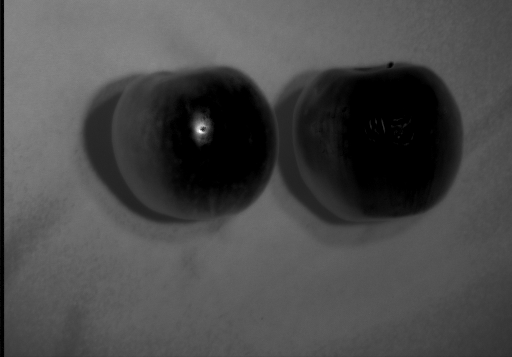}}
\hfill
	\subfloat[]{\papergraphic[width=0.135\textwidth]{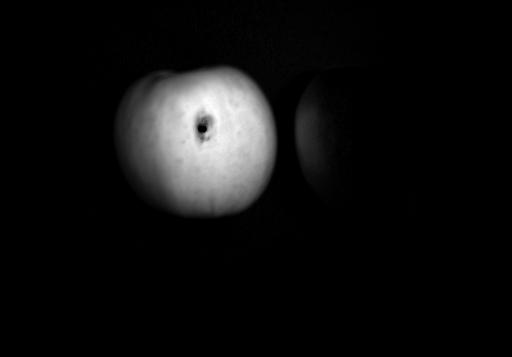}}
\hfill
	\subfloat[]{\papergraphic[width=0.135\textwidth]{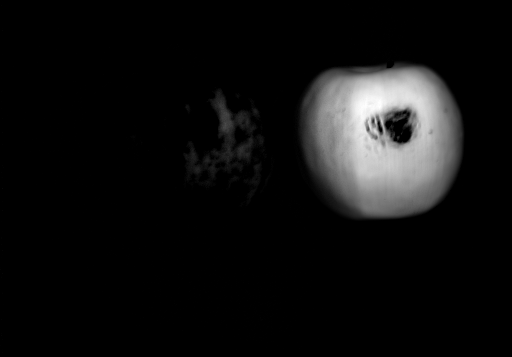}}
\hfill
	\subfloat[]{\papergraphic[width=0.135\textwidth]{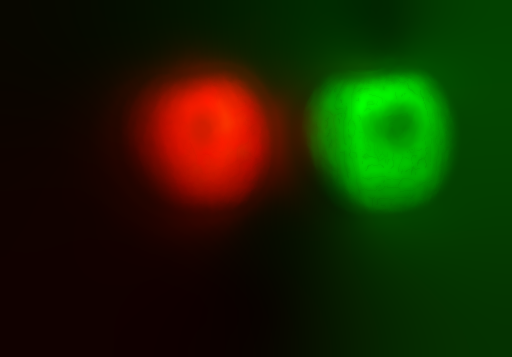}}
\hfill
	\subfloat[]{\papergraphic[width=0.135\textwidth]{Figures/chapter5/results/apple/boundingBox2.png}}
	\caption{Original exploratory result for objects at similar depth with different materials. (a) Band image; (b) sparse optical cues; (c)--(e) material maps; (f) propagated scores; (g) bounding boxes.}
\label{fig:ocres2} 
\end{figure*}

Figure~\ref{fig:ocres2} examines two objects at similar depth but with different material composition. The optical responses do not provide a strong, independently verified depth distinction between the apples, whereas the material maps provide contrasting responses over their visible surfaces. The resulting spatial separation is therefore consistent with the value of spectral information when depth contrast is weak. A bright response within an apple also indicates sensitivity to surface reflection: an interior optical response should not automatically be interpreted as a separate depth layer.

The experiment illustrates a useful operating condition for cue fusion, but it does not prove that the optical branch improves this particular scene. Both material-only processing and the complete pipeline would need to be evaluated on the same annotations to establish that incremental benefit. The displayed red and green regions encode different output scores; they are not predicted fruit classes.

\begin{figure*}[!t]
	\centering
	\subfloat[]{\papergraphic[width=0.135\textwidth]{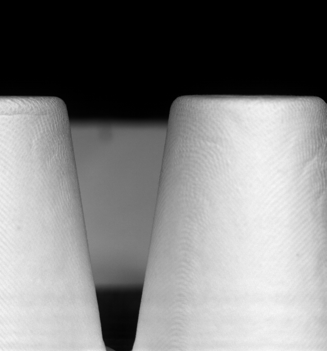}}
\hfill
	\subfloat[]{\papergraphic[width=0.135\textwidth]{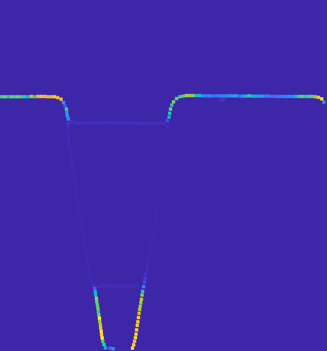}}
\hfill
	\subfloat[]{\papergraphic[width=0.135\textwidth]{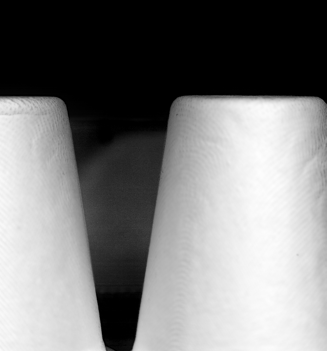}}
\hfill
	\subfloat[]{\papergraphic[width=0.135\textwidth]{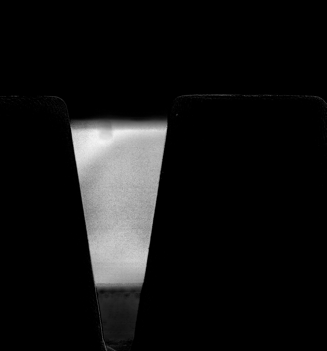}}
\hfill
	\subfloat[]{\papergraphic[width=0.135\textwidth]{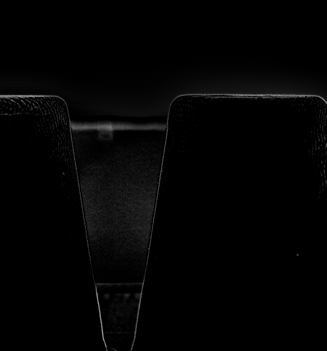}}
\hfill
	\subfloat[]{\papergraphic[width=0.135\textwidth]{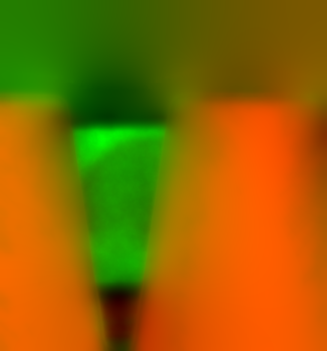}}
\hfill
	\subfloat[]{\papergraphic[width=0.135\textwidth]{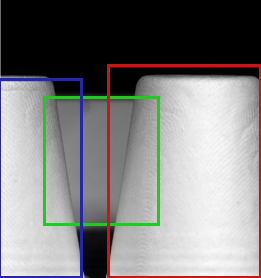}}		
	\caption{Original exploratory result for objects at similar depth and with the same material. (a) Band image; (b) sparse optical cues; (c)--(e) material maps; (f) propagated scores; (g) bounding boxes. Instance separation remains difficult in this setting.}
\label{fig:ocres3} 
\end{figure*}

Figure~\ref{fig:ocres3} provides a complementary stress case: two cups share similar material and depth. Their propagated values overlap, leaving a weak basis for separating the instances by a scalar score alone. Distinct boxes can still arise from spatial boundaries and connected-region processing, but such a result depends on the postprocessing rule. If the exposed regions became connected, or if an occlusion removed the separating contour, the same evidence could support a merged region. This example identifies a limitation of representing object identity through a single continuous field, rather than demonstrating robust instance separation under both ambiguities.

\subsection{Different Depths and Outdoor Scenes}
\begin{figure*}[!t]
	\centering
	\subfloat[]{\papergraphic[width=0.19\textwidth]{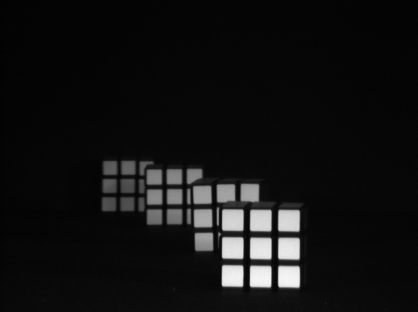}}
\hfill
	\subfloat[]{\papergraphic[width=0.19\textwidth]{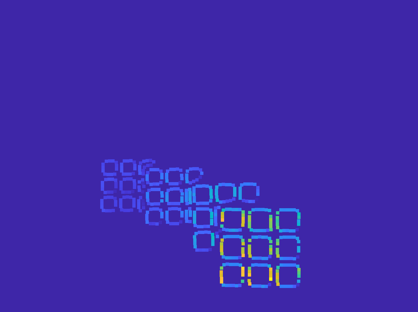}}
\hfill
	\subfloat[]{\papergraphic[width=0.19\textwidth]{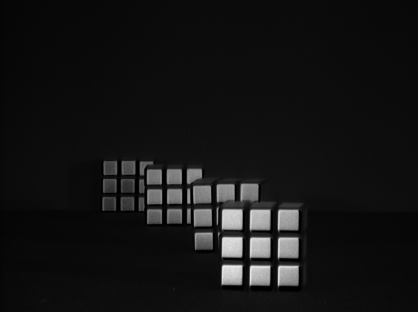}}
\hfill
	\subfloat[]{\papergraphic[width=0.19\textwidth]{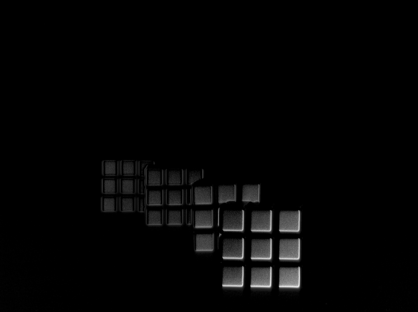}}
\hfill
	\subfloat[]{\papergraphic[width=0.19\textwidth]{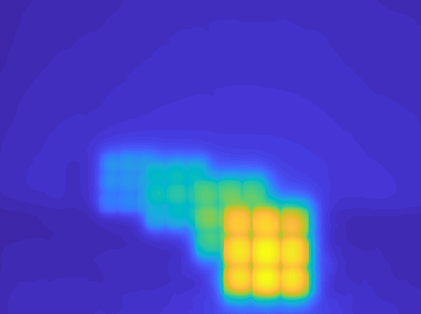}}
	\\
	\subfloat[]{\papergraphic[width=0.19\textwidth]{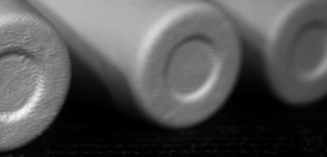}}
\hfill
	\subfloat[]{\papergraphic[width=0.19\textwidth]{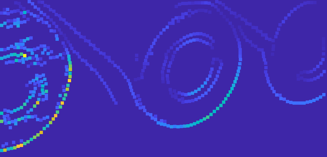}}
\hfill
	\subfloat[]{\papergraphic[width=0.19\textwidth]{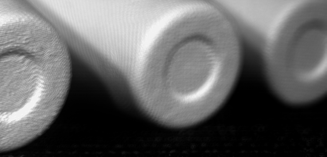}}
\hfill
	\subfloat[]{\papergraphic[width=0.19\textwidth]{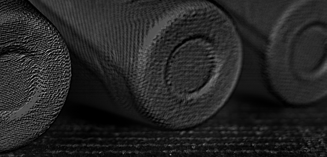}}
\hfill
	\subfloat[]{\papergraphic[width=0.19\textwidth]{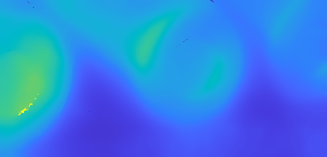}}		
	\caption{Original exploratory results for objects with similar material and different depths. (a), (f) band images; (b), (g) sparse optical cues; (c)--(d), (h)--(i) material maps; (e), (j) propagated scores. These examples do not include a bounding-box evaluation.}
\label{fig:ocres4} 
\end{figure*}

Figure~\ref{fig:ocres4} examines the converse setting: similar material appearance with different depth arrangements. In the repeated-block scene, a material map can respond to several exposed structures simultaneously, while the sparse optical map supplies location-dependent responses around their boundaries. The propagated field assigns different values across parts of the arrangement. This is consistent with optical cues helping to distinguish surfaces that material grouping alone might join, although the colours do not establish measured depth ordering or error.

The cup scene highlights an additional interaction between the cues. Saturated regions lose useful spectral variation and can also distort an optical score computed from the same measurements. The two branches are therefore complementary but not statistically independent: one acquisition defect can affect both. Smooth propagation can extend an erroneous seed into a larger region when the graph strongly couples neighbouring pixels. Neither scene reports bounding-box performance, so these panels support analysis of the intermediate representation rather than a complete detection claim.

\begin{figure*}[!t]
\centering

		\subfloat[]{\papergraphic[width=0.135\textwidth]{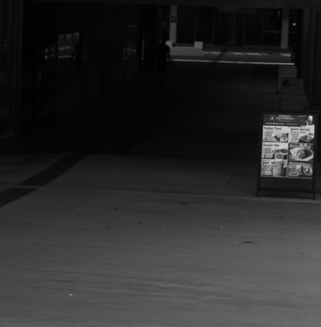}}
\hfill
		\subfloat[]{\papergraphic[width=0.135\textwidth]{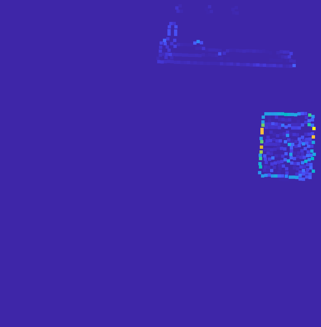}}
\hfill
		\subfloat[]{\papergraphic[width=0.135\textwidth]{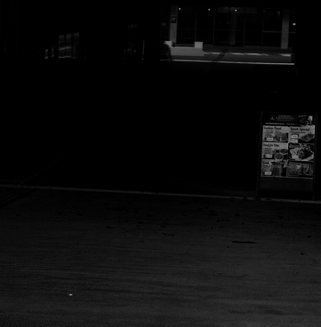}}
\hfill
		\subfloat[]{\papergraphic[width=0.135\textwidth]{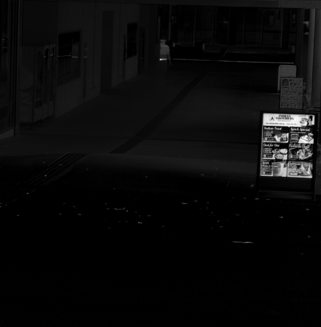}}
\hfill
		\subfloat[]{\papergraphic[width=0.135\textwidth]{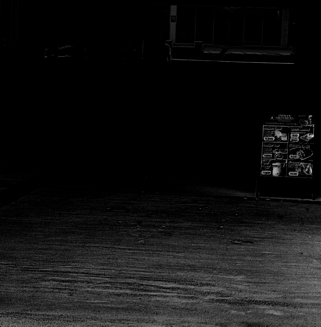}}
\hfill
		\subfloat[]{\papergraphic[width=0.135\textwidth]{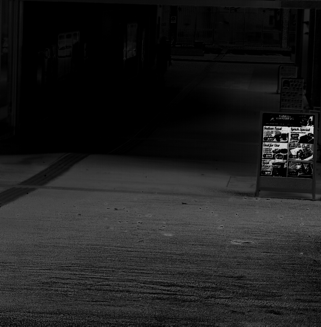}}
\hfill
		\subfloat[]{\papergraphic[width=0.135\textwidth]{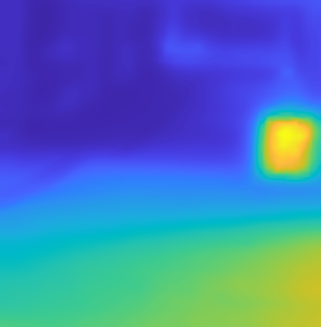}}					
	\caption{Original exploratory outdoor result. (a) Band image; (b) sparse optical cues; (c)--(f) material maps; (g) propagated scores. No detection boxes or quantitative depth errors are reported.}
\label{fig:ocres5} 
\end{figure*}

Figure~\ref{fig:ocres5} presents an outdoor scene with overlapping structures and heterogeneous surfaces. The sparse responses cover only part of the scene, whereas the final field contains broad, smoothly varying regions. The dense appearance is consequently evidence of propagation, not of a dense independent depth measurement. Some material maps respond across several structural elements, illustrating again that a material component is not equivalent to an object instance.

Outdoor illumination and surface-dependent reflectance make interpretation especially difficult. An apparent score change can reflect a material boundary, a brightness difference, an optical effect, or their combination. Without independent depth measurements and common display scaling, the final colours cannot verify geometric ordering or be compared numerically with colours in other scenes. Thin structures and partially exposed surfaces also test whether the neighbourhood graph preserves useful boundaries.

\FloatBarrier
\section{Limitations}\label{sec:limitations}
Material identity is not object identity: one object may contain several materials, and several objects may share one material. Similarly, optical variation is influenced by depth, texture, illumination, reflectance, saturation, noise, and camera calibration. When material and optical evidence are both ambiguous, graph smoothing can merge instances or spread an incorrect seed. A single view provides no direct evidence of a fully occluded surface. The present observations therefore support investigating cue complementarity, not a general solution to occlusion.

\Needspace{12\baselineskip}
\section{Conclusion}\label{sec:ch5-conclusion}
This exploratory study considers material and wavelength-dependent optical cues for localising partially occluded objects from one hyperspectral acquisition. The original qualitative examples suggest complementary information in selected scenes and expose limitations when objects share both material and depth. The formulation builds on established clustering and spectral-depth components and makes their mathematical assumptions explicit. Quantitative validation, reproducible implementation, and independent depth or instance annotations remain necessary before broader performance claims can be made.

\bibliographystyle{IEEEtran}
\bibliography{occ_bib_settings,references_occ_v2}

\begin{thebibliography}{10}
\providecommand{\url}[1]{#1}
\csname url@samestyle\endcsname
\providecommand{\newblock}{\relax}
\providecommand{\bibinfo}[2]{#2}
\providecommand{\BIBentrySTDinterwordspacing}{\spaceskip=0pt\relax}
\providecommand{\BIBentryALTinterwordstretchfactor}{4}
\providecommand{\BIBentryALTinterwordspacing}{\spaceskip=\fontdimen2\font plus
\BIBentryALTinterwordstretchfactor\fontdimen3\font minus \fontdimen4\font\relax}
\providecommand{\BIBforeignlanguage}[2]{{%
\expandafter\ifx\csname l@#1\endcsname\relax
\typeout{** WARNING: IEEEtran.bst: No hyphenation pattern has been}%
\typeout{** loaded for the language `#1'. Using the pattern for}%
\typeout{** the default language instead.}%
\else
\language=\csname l@#1\endcsname
\fi
#2}}
\providecommand{\BIBdecl}{\relax}
\BIBdecl

\bibitem{xu2013efficient}
Y.~Xu, L.~Qin, G.~Li, and Q.~Huang, ``An efficient occlusion detection method to improve object trackers,'' in \emph{IEEE International Conference on Image Processing}, 2013, pp. 2445--2449.

\bibitem{ghiasi2014occlusion}
G.~Ghiasi and C.~C. Fowlkes, ``Occlusion coherence: {L}ocalizing occluded faces with a hierarchical deformable part model,'' in \emph{{IEEE} Conference on Computer Vision and Pattern Recognition}, 2014, pp. 2385--2392.

\bibitem{burgos2013robust}
X.~P. Burgos-Artizzu, P.~Perona, and P.~Doll{\'a}r, ``Robust face landmark estimation under occlusion,'' in \emph{{IEEE} International Conference on Computer Vision}, 2013, pp. 1513--1520.

\bibitem{noh2018improving}
J.~Noh, S.~Lee, B.~Kim, and G.~Kim, ``Improving occlusion and hard negative handling for single-stage pedestrian detectors,'' in \emph{IEEE Conference on Computer Vision and Pattern Recognition}, 2018, pp. 966--974.

\bibitem{Oberweger2018}
M.~Oberweger, M.~Rad, and V.~Lepetit, ``Making deep heatmaps robust to partial occlusions for {3D} object pose estimation,'' in \emph{The European Conference on Computer Vision}, 2018.

\bibitem{Lowe2004}
D.~Lowe, ``Distinctive image features from scale-invariant keypoints,'' \emph{International Journal of Computer Vision}, vol.~60, no.~2, pp. 91--110, 2004.

\bibitem{Girshick2011}
R.~B. Girshick, P.~F. Felzenszwalb, and D.~McAllester, ``Object detection with grammar models,'' in \emph{Proceedings of the 24th International Conference on Neural Information Processing Systems}, ser. NIPS’11, 2011, p. 442–450.

\bibitem{Felzenszwalb2008}
P.~Felzenszwalb, D.~McAllester, and D.~Ramanan, ``A discriminatively trained, multiscale, deformable part model,'' in \emph{{IEEE} Conference on Computer Vision and Pattern Recognition}, 2008.

\bibitem{Meger2011}
D.~Meger, C.~Wojek, J.~Little, and B.~Schiele, ``Explicit occlusion reasoning for {3D} object detection,'' in \emph{Proceedings of the British Machine Vision Conference 2011}, 2011.

\bibitem{Shu2012}
G.~Shu, A.~Dehghan, O.~Oreifej, E.~Hand, and M.~Shah, ``Part-based multiple-person tracking with partial occlusion handling,'' in \emph{{IEEE} Conference on Computer Vision and Pattern Recognition}, 2012.

\bibitem{Wang2013}
T.~Wang, X.~He, and N.~Barnes, ``Learning structured hough voting for joint object detection and occlusion reasoning,'' in \emph{{IEEE} Conference on Computer Vision and Pattern Recognition}, 2013.

\bibitem{Zia2015}
A.~Zia, J.~Zhou, and Y.~Gao, ``Relative depth estimation from hyperspectral data,'' in \emph{International Conference on Digital Image Computing: Techniques and Applications}.\hskip 1em plus 0.5em minus 0.4em\relax {IEEE}, Nov. 2015.

\bibitem{Zia2021TIP}
A.~Zia, J.~Zhou, and Y.~Gao, ``Exploring chromatic aberration and defocus blur for relative depth estimation from monocular hyperspectral image,'' \emph{IEEE Transactions on Image Processing}, vol.~30, pp. 4357--4370, 2021.

\bibitem{Zia2020Thesis}
\BIBentryALTinterwordspacing
A.~Zia, ``Exploring extrinsic spectral properties of hyperspectral image for {3D} computer vision,'' Ph.D. dissertation, Griffith University, 2020. [Online]. Available: \url{https://research-repository.griffith.edu.au/items/6c4ccabf-f952-4780-8592-ac6f8ea62dae}
\BIBentrySTDinterwordspacing

\bibitem{Hanson2022Occlusion}
N.~Hanson, G.~Lvov, and T.~Padir, ``Occluded object detection and exposure in cluttered environments with automated hyperspectral anomaly detection,'' \emph{Frontiers in Robotics and AI}, vol.~9, p. 982131, 2022.

\bibitem{Sun2025Grasp}
\BIBentryALTinterwordspacing
Z.~Sun, Z.~Dong, S.~Wang, Z.~Chu, and F.~Chen, ``A hyperspectral imaging guided robotic grasping system,'' \emph{IEEE Robotics and Automation Letters}, 2025. [Online]. Available: \url{https://arxiv.org/abs/2512.05578}
\BIBentrySTDinterwordspacing

\bibitem{Gillis2015}
N.~Gillis, D.~Kuang, and H.~Park, ``Hierarchical clustering of hyperspectral images using rank-two nonnegative matrix factorization,'' \emph{{IEEE} Transactions on Geoscience and Remote Sensing}, vol.~53, no.~4, pp. 2066--2078, 2015.

\bibitem{Qian2011}
Y.~Qian, S.~Jia, J.~Zhou, and A.~Robles-Kelly, ``Hyperspectral unmixing via {L-1/2} sparsity-constrained nonnegative matrix factorization,'' \emph{{IEEE} Transactions on Geoscience and Remote Sensing}, vol.~49, no.~11, pp. 4282--4297, 2011.

\bibitem{tong2016nonnegative}
L.~Tong, J.~Zhou, Y.~Qian, X.~Bai, and Y.~Gao, ``Nonnegative matrix factorization based hyperspectral unmixing with partially known endmembers,'' \emph{{IEEE} Transactions on Geoscience and Remote Sensing}, vol.~54, no.~11, pp. 6531--6544, 2016.

\bibitem{pentland1987}
A.~P. Pentland, ``{A New Sense for Depth of Field},'' \emph{IEEE Transactions on Pattern Analysis and Machine Intelligence}, vol.~9, no.~4, pp. 523--531, 1987.

\bibitem{Persch2017}
N.~Persch, C.~Schroers, S.~Setzer, and J.~Weickert, ``Physically inspired depth-from-defocus,'' \emph{Image and Vision Computing}, vol.~57, pp. 114--129, 2017.

\bibitem{Garcia2000}
J.~Garcia, J.~Sanchez, X.~Orriols, and X.~Binefa, ``Chromatic aberration and depth extraction,'' in \emph{Proceedings 15th International Conference on Pattern Recognition. {ICPR}-2000}, vol.~1.\hskip 1em plus 0.5em minus 0.4em\relax {IEEE} Comput. Soc, 2000, pp. 762--765.

\bibitem{trouve2013}
P.~Trouv\'{e}, F.~Champagnat, G.~L. Besnerais, J.~Sabater, T.~Avignon, and J.~Idier, ``Passive depth estimation using chromatic aberration and a depth from defocus approach,'' \emph{Applied Optics}, vol.~52, no.~29, pp. 7152--7164, Oct. 2013.

\bibitem{Gillis2014}
N.~Gillis and S.~A. Vavasis, ``Fast and robust recursive algorithms for separable nonnegative matrix factorization,'' \emph{{IEEE} Transactions on Pattern Analysis and Machine Intelligence}, vol.~36, no.~4, pp. 698--714, 2014.

\bibitem{Arora2016}
S.~Arora, R.~Ge, R.~Kannan, and A.~Moitra, ``Computing a nonnegative matrix factorization---provably,'' \emph{{SIAM} Journal on Computing}, vol.~45, no.~4, pp. 1582--1611, 2016.

\bibitem{Kullback1951}
S.~Kullback and R.~A. Leibler, ``On information and sufficiency,'' \emph{The Annals of Mathematical Statistics}, vol.~22, no.~1, pp. 79--86, 1951.

\bibitem{Grady2006}
L.~Grady, ``Random walks for image segmentation,'' \emph{{IEEE} Transactions on Pattern Analysis and Machine Intelligence}, vol.~28, no.~11, pp. 1768--1783, 2006.

\bibitem{Duchenne2008}
O.~Duchenne, J.-Y. Audibert, R.~Keriven, J.~Ponce, and F.~Segonne, ``Segmentation by transduction,'' in \emph{{IEEE} Conference on Computer Vision and Pattern Recognition}.\hskip 1em plus 0.5em minus 0.4em\relax {IEEE}, 2008.

\bibitem{Levin2008}
A.~Levin, D.~Lischinski, and Y.~Weiss, ``{A Closed-Form Solution to Natural Image Matting},'' \emph{{IEEE} Transactions on Pattern Analysis and Machine Intelligence}, vol.~30, no.~2, pp. 228--242, Feb. 2008.

\bibitem{Singaraju2009}
D.~Singaraju, L.~Grady, and R.~Vidal, ``P-brush: {C}ontinuous valued {MRFs} with normed pairwise distributions for image segmentation,'' in \emph{{IEEE} Conference on Computer Vision and Pattern Recognition}, 2009.

\end{thebibliography}

\end{document}